\documentclass[10pt,twocolumn,letterpaper]{article}

\usepackage{cvpr}              

\usepackage{graphicx}
\usepackage{amsmath}
\usepackage{amssymb}
\usepackage{booktabs}
\usepackage{multirow,multicol}
\usepackage{pifont}

\usepackage{algorithm}
\usepackage{algorithmicx}
\usepackage{algpseudocode}

\usepackage[dvipsnames]{xcolor}
\usepackage{color}
\definecolor{myGreen}{RGB}{55,149,73}

\newcommand{\cmark}{\textcolor{myGreen}{\ding{51}}}
\usepackage[pagebackref,breaklinks,colorlinks]{hyperref}

\usepackage[capitalize]{cleveref}
\crefname{section}{Sec.}{Secs.}
\Crefname{section}{Section}{Sections}
\Crefname{table}{Table}{Tables}
\crefname{table}{Tab.}{Tabs.}

\newcommand{\figref}[1]{Figure \ref{#1}}

\newcommand{\name}{CD\textsuperscript{2}-pFed}
\newcommand{\pfl}{PFL}

\newcommand{\para}[1]{\vspace{.05in}\noindent\textbf{#1}}

\def\cvprPaperID{7502} 
\def\confName{CVPR}
\def\confYear{2022}

\begin{document}


\title{CD$^2$-pFed: Cyclic Distillation-guided Channel Decoupling for Model Personalization in Federated Learning}

\author{
Yiqing Shen\textsuperscript{1}, 
Yuyin Zhou\textsuperscript{2}, 
Lequan Yu\textsuperscript{3}\thanks{Corresponding Author.}\\
\textsuperscript{1} Shanghai Jiao Tong University,
\textsuperscript{2} UC Santa Cruz,
\textsuperscript{3} The University of Hong Kong 
\\
{\tt\small shenyq@sjtu.edu.cn, zhouyuyiner@gmail.com, lqyu@hku.hk}
}
\maketitle

\begin{abstract}
Federated learning (FL) is a distributed learning paradigm that enables multiple clients to collaboratively learn a shared global model. Despite the recent progress, it remains challenging to deal with heterogeneous data clients, as the discrepant data distributions usually prevent the global model from delivering good generalization ability on each participating client. In this paper, we propose \textbf{\name}, a novel \textbf{C}yclic \textbf{D}istillation-guided \textbf{C}hannel \textbf{D}ecoupling framework, to personalize the global model in FL, under various settings of data heterogeneity. Different from previous works which establish layer-wise personalization to overcome the non-IID data across different clients, we make the first attempt at channel-wise assignment for model personalization, referred to as channel decoupling. To further facilitate the collaboration between private and shared weights, we propose a novel cyclic distillation scheme to impose a consistent regularization between the local and global model representations during the federation. Guided by the cyclical distillation, our channel decoupling framework can deliver more accurate and generalized results for different kinds of heterogeneity, such as feature skew, label distribution skew, and concept shift. Comprehensive experiments on four benchmarks, including natural image and medical image analysis tasks, demonstrate the consistent effectiveness of our method on both local and external validations.  
\end{abstract}
\section{Introduction}
Deep learning techniques have received notable attention in various vision tasks, such as image classification~\cite{he2016deep}, object detection~\cite{ren2015faster}, and semantic segmentation~\cite{long2015fully}.
Yet, the success of deep neural networks heavily relies on a tremendous volume of valuable training images. One possible solution is to collaboratively curate numerous data samples from different parties (\eg, different mobile devices and companies).
However, collecting distributed data into a centralized storage facility is costly and time-consuming.
Additionally, in real practice, decentralized image data should not be directly shared, due to privacy concerns or legal restrictions~\cite{gdpr,hippa}.
In this case, conventional centralized machine learning frameworks fail to satisfy the data privacy protection constraint. 

\begin{figure}[t]
\centering
\includegraphics[width=1.0\linewidth]{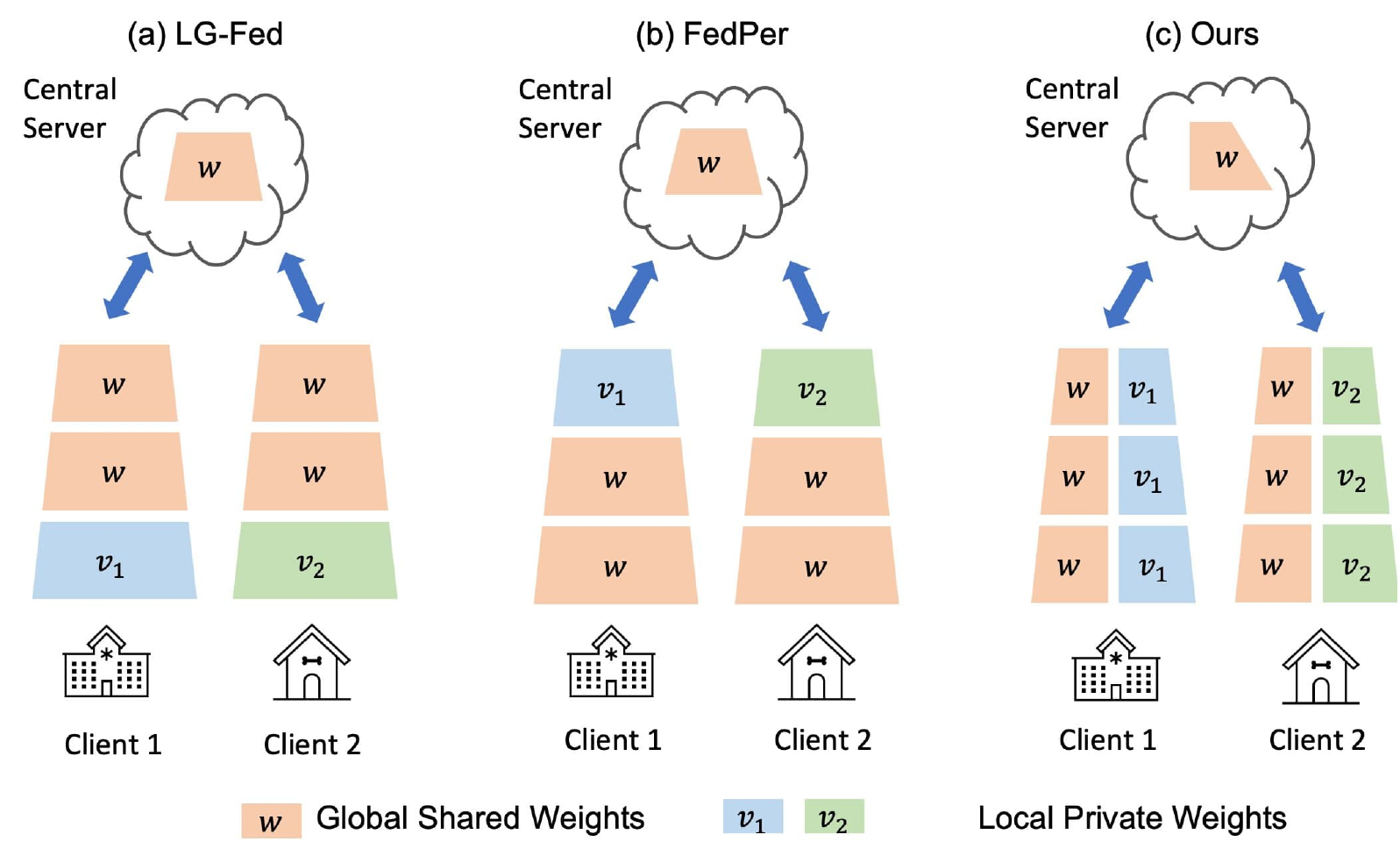}
\caption{Illustration of different parameter decoupling manners for model personalization in Federated Learning. The previous approaches combine local and global parameters in a layer-wise mechanism, including LG-Fed~\cite{lgfed} in low-level input layers (a) and FedPer~\cite{fedper} in high-level output layers (b). 
Instead, we achieve model personalization via channel-wise decoupling (c).}
\label{fig:demo}
\end{figure}

Therefore, the data-private distributed training paradigms, especially Federated Learning (FL), have received an increasing popularity~\cite{fedavg,fl,Sun_2021_CVPR,Zhuang_2021_ICCV,Zhang_2021_ICCV,Li_2021_CVPR,Liu_2021_CVPR,Gong_2021_ICCV,Guo_2021_CVPR}. 
To be more specific, in FL, a shared model is globally trained with an orchestration of local updates within data stored at each client. 
A pioneering FL algorithm named Federated Average (FedAVG), aggregates parameters at the central server by communication across clients once per global epoch, without explicit data sharing~\cite{fedavg}. 
Compared with local training, the federation on a larger scale of training data has demonstrated its superiority to boost the generalization ability on unseen data, with the orchestration of distributed private data~\cite{fedavg,fedmixed}.

However, data heterogeneity is one of the most fundamental challenges faced by FL. 
The concept of independent and identically distributed (IID) is clear, while data can be non-IID in many ways, \eg, feature skew, label distribution skew, or concept shift~\cite{openchallenge}.
Previously, sharp performance degradation was observed on FedAVG with unbalanced and non-IID data. 
This ill-effect is attributed to the weight divergence, which can be quantified by the earth mover’s distance between distributions over classes \cite{fednoniid}. 
Although each client can train a private model locally by optimizing the objective with no information change among each other, it would inevitably result in overfitting and a poor generalization ability on new samples. 
As suggested in~\cite{fedavg}, simply sharing a small subset of data globally greatly enhances the generalization of FedAVG.
However, this scheme cannot be directly applied to real-world tasks due to the violation of privacy concerns.

Consequently, researchers have sought to train a collection of models that is stylized for each local distribution to enable stronger performance for each participating client without requiring any data sharing~\cite{fednoniid}, which is known as personalized federated learning \pfl~\cite{tan2021towards}.
Various approaches have been proposed to accomplish the model personalization in FL~\cite{tan2021towards,meta,multi_task,fedmixed}.
Among these different paradigms, one popular solution is to directly assign personalized parameters for each local client.
For this line of methods, the private personalized parameters are trained locally and not shared with the central server.
Existing works have made attempts to achieve personalization by assigning personalized parameters in either top layers~\cite{fedper} or bottom layers~\cite{lgfed}. 
However, these approaches usually require prior knowledge for the determination of which layers to be personalized. 
More critically, we observe performance degradation that existing \pfl~approaches fail to achieve a consistent generalization over comprehensive settings of data heterogeneity~\cite{quinonero2009dataset}.
Additionally, existing layer-wise personalization approaches cannot effectively handle the discrepancy between the learned local and global model representations due to the weight divergence \cite{fednoniid}. 
The inferior performance of some local clients motivates us to seek a more generic yet efficient combination between the local and global information.

In light of these challenges, we propose \textbf{\name}, a novel Cyclic Distillation-guided Channel Decoupling framework for model personalization in FL.
As shown in \figref{fig:demo}, different from previous layer-wise personalization approaches, \eg, FedPer~\cite{fedper} and LG-Fed~\cite{lgfed}, the proposed novel \emph{channel decoupling} paradigm dynamically decouples the parameters at the channel dimension for personalization instead. 
By employing learnable personalized weights at all layers, our channel decoupling paradigm no longer requires heuristics for designing specific personalization layers.
More importantly, our method achieves model personalization for both low-level and high-level layers, which facilitates tackling feature heterogeneity, distribution skew, and concept shift.

To bridge the semantic gap between the learned visual representation from the decoupled channels, we further propose a novel \emph{cyclic distillation} scheme by mutually distilling the local and global model representation (\ie, soft predictions by the private and shared weights) from each other.
Benefiting from the distilled knowledge, our channel decoupling framework enables synergistic information exchange between the global and local model training, therefore preventing biased local model training on non-IID data.
Extensive experimental results on both heterogeneous data and exterior unseen samples~\cite{lgfed} demonstrate that our method largely improves the generalization of FedAVG with negligible additional computation overhead.
Below, we summarize the major contributions of this work.
\begin{itemize}
    \item We propose a novel \emph{channel decoupling} paradigm to decouple the global model at the channel dimension for personalization. 
    Instead of using personalization layers for tackling either feature or label distribution skew, our approach provides a unified solution to address a broad range of data heterogeneity.

    \item To further enhance the collaboration between private and shared weights in channel decoupling, we design a novel \emph{cyclic distillation} scheme to narrow the divergence between them.

    \item We compare our method with previous state-of-the-art \pfl~approaches on four benchmark datasets, including synthesized and real-world image classification tasks, with different kinds of heterogeneity. Results demonstrate the superiority of our method over state-of-the-art \pfl~approaches.
    
\end{itemize}

\begin{figure*}[!htb]
\centering
\includegraphics[width=0.865\linewidth]{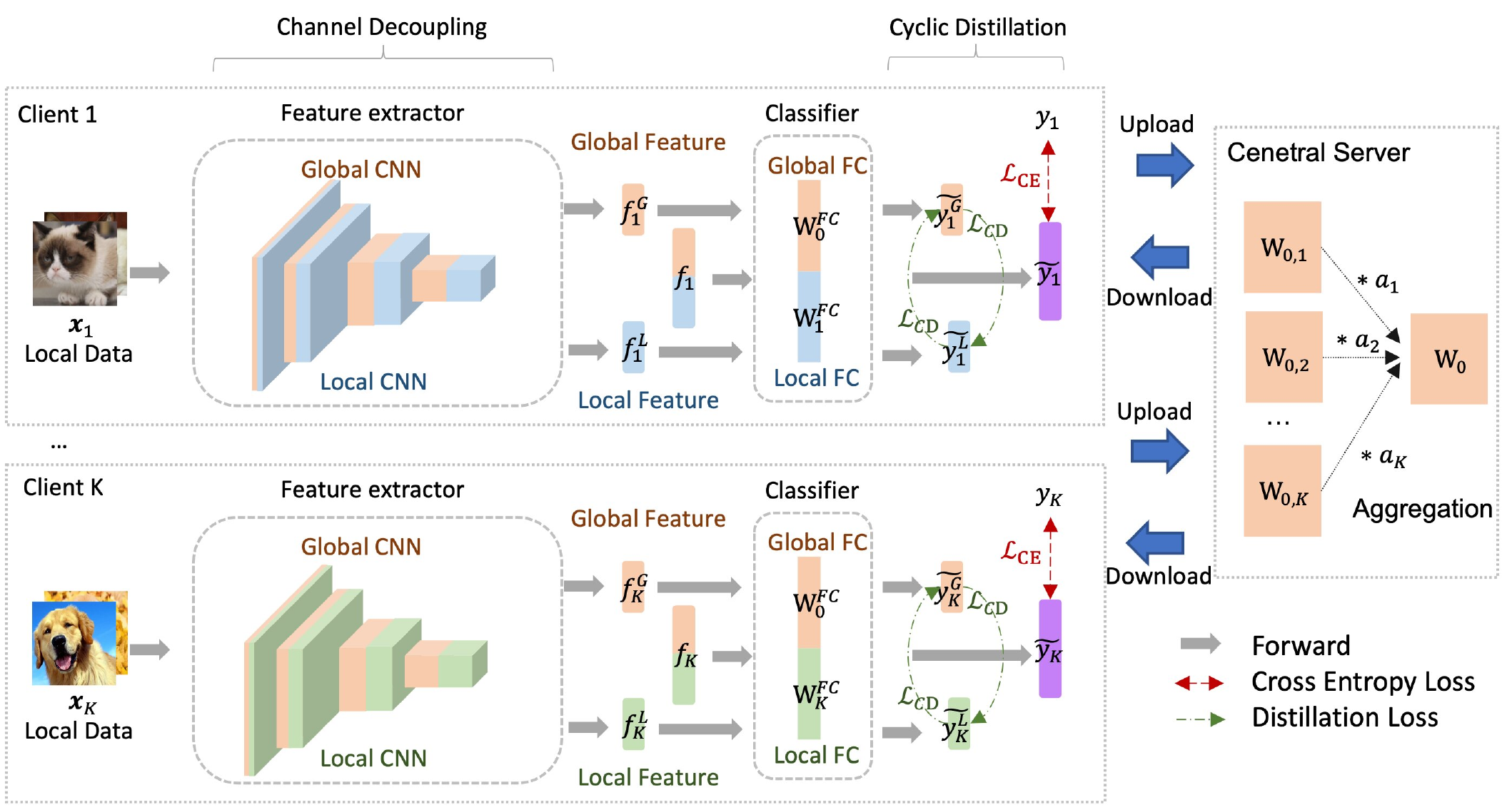}
\caption{A schematic illustration of our proposed \name~framework for model personalization in federated learning. We use the blue and green to mark out personalized channels and features, which reside locally; and the orange for global representations.}
\label{fig:framework}
\end{figure*}

\section{Related Work}
\subsection{Data Heterogeneity}
Federated Averaging (FedAVG) is a prevailing federated learning algorithm to train a global model with distributed data \cite{fedavg}. 
Under the assumption of unbalanced and IID (independent and identically distributed) characteristics, FedAvg has achieved notable empirical success on its robustness and performance. 
However, facing the variety and diversity of real-world data, the condition of non-IID is unrealistic to be ensured. 
Instead, statistical data heterogeneity is a more general case, where the data distribution of local clients deviates significantly from the global distribution. It results in sharp performance degradation, \eg, up to 11.0\% for MNIST and 51.0\% for CIFAR-10 \cite{fednoniid}, where the common predictor does not generalize well on local data. 
One of the most fundamental challenges in training a robust FL is the presence of non-IID data. 
Concretely, the underlying distribution for arbitrary paired clients is very likely to differ from each other. There are various formulations on the existence of non-IID, including the feature distribution skew, label distribution skew, and the concept shift \cite{openchallenge}. 
To tackle the data heterogeneity, Li \etal proposed an optimization scheme, namely FedProx, to re-parametrize FedAvg with variable amounts of work to be performed locally across devices \cite{fedprox}. FedBN employs local batch normalization to alleviate the feature distribution skew before the model aggregation \cite{fedbn}.

\subsection{Personalized Federated Learning}
Three major challenges that restrict the generalization ability of federated global model on local data are 1) device heterogeneity, 2) data heterogeneity due to the non-IID distribution, 3) model heterogeneity to adapt to local environment \cite{whypfed}. Among those, data-IID is the most practical issue. Yet, the privacy protection mechanisms in conventional FL conflict with achieving higher performance for each individual user \cite{whypfed2}. Subsequently, Personalized Federated Learning (PFL) has received numerous attentions from researchers to cope with the above three challenges. In PFL, the global model is personalized for each local client and plays an intermediate paradigm between pure local training and FL \cite{pfl}. Leveraging a personalized model per client, PFL can integrate the client’s own dataset and orchestration of data from other clients into the training process. 

There are various techniques to adapt the global model for personalization \cite{survey}, including transfer learning \cite{transfer}, multi-task learning \cite{multi_task}, meta-learning \cite{meta}, knowledge distillation \cite{distill}, and network decoupling \cite{fedper,lgfed}.
We mainly focus on the network decoupling methods, where the global network is decoupled into personalized layers which reside locally, and global layers. For example, FedPer splits a neural network architecture into base layers, which are trained centrally by FedAvg, and the top personalized layers \cite{fedper}, which are trained separately with the personalized layers. Very similarly to FedPer, LG-Fed personalizes the bottom layer, while keeping the top layers shared across all involved clients \cite{lgfed}. FedPer shows its superiority with an observable labeling skew such as on FLICKR-AES \cite{pia}, while LG-Fed with data skews such as CIFAR non-IID split. However, both skews exist in real-world tasks. Therefore, in this work, we attempt to bridge the gap between top layers personalization and bottom layers personalization with a unified channel decoupling scheme.

\subsection{Knowledge Distillation}
The key idea of knowledge distillation (KD) is to transfer the dark knowledge from a pre-trained teacher model to a lightweight student network by learning its soft predictions, intermediate features or attention maps~\cite{kd,quinonero2009dataset,kd3}. 
KD has broad applications in machine learning and computer visions fields, including transfer learning, semi-supervised learning, reinforcement learning \cite{kd0,kd2,kd4,zhang2020distilling,xu2020knowledge}.
KD has also achieved remarkable performance as a regularization scheme. 
For example, Yun~\etal~\cite{cskd} distilled predictive distribution between samples of the same label to mitigate overconfident predictions and reduce intra-class variations in an image classification task.%
In FL scenarios, an ensemble distillation scheme was proposed to replace the aggregation for model fusion~\cite{kdfl}. 
Besides, Li~\etal~\cite{fedmd} adopted knowledge distillation to personalize the global model by translating knowledge between participants.

\section{Methodology}
\subsection{Problem Formulation}
We consider a set of $K$ clients, which are all connected to a central server. 
Moreover, each client only has access to its local data, denoted as $\mathcal{D}_i$, with no data sharing between clients. 
In a data heterogeneous setting, the underlying distribution of $\mathcal{D}_i$, denoted as $\mathcal{P}_i$ are not identical, \ie, $\mathcal{P}_i \not= \mathcal{P}_j$. Specifically, there are three common categories to depict the non-IID characteristic in FL~\cite{openchallenge}: 1) feature distribution skew (covariate shift), \ie, $\mathcal{P}_i(x) \not= \mathcal{P}_j(x)$; 2) label distribution skew (prior probability shift), \ie, $\mathcal{P}_i(y) \not= \mathcal{P}_j(y)$; and 3) same label but different features (concept drift), \ie, $\mathcal{P}_i(x|y) \not= \mathcal{P}_j(x|y)$.

The goal of our work is to train a collection of $K$ models to adapt to the local dataset without exchanging their local data with other parties. 
The network at the $i$-th client ($i\in\{1,\cdots,K\}$) is composed of private personalized parameters $w_i$, and global shared parameters $w_0$. 
More formally, we formulate the loss function corresponding to the $i$-th client as $F_i:\mathbb{R}^d\to\mathbb{R}$, and then the overall objective in personalized federated learning is defined as follows,
\begin{equation}
    \min_{\{w_i\}_{i=0}^K} F(w_0,w_1,\cdots,w_K) = \sum_{i=1}^K \alpha_i \cdot F_i(w_0,w_K).
\end{equation}
The balancing weight $\alpha_i$ depends on the scale of the private dataset, \ie, $\alpha_i = \frac{|\mathcal{D}_i|}{\sum_{j} |\mathcal{D}_j|}$.
In this scenario, we consider supervised learning, leading to 
\begin{equation}
F_i(w_0,w_i) = \mathbb{E}_{(\textbf{x}_j,y_j) \sim \mathcal{P}_i} \left[ l_i(\textbf{x}_j,y_j;w_0,w_i)\right],
\label{eq:local_loss}
\end{equation}
where $l_i$ measures the sample-wise loss between the prediction of the network parameterized by $(w_0, w_i)$ and the ground truth label $y_j$ when given the input image $\textbf{x}_i$.

\begin{figure}[!t]
\centering
\includegraphics[width=0.8\linewidth]{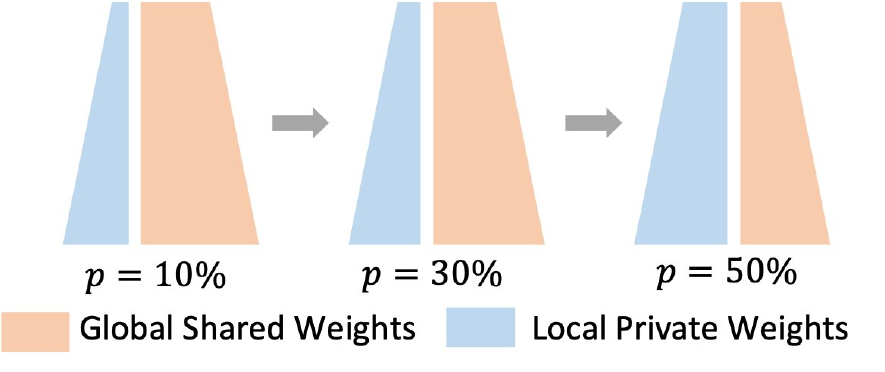}
\caption{Illustration of Channel Decoupling with progressive model personalization.}
\label{fig:kd}
\end{figure}

\subsection{Channel Decoupling for Model Personalization}
As shown in Figure~\ref{fig:framework}, we propose a vertically channel-wise decoupling framework to personalize the global model for non-IID federated learning. 
Concretely, we assign an adaptive proportion of learnable personalized weights at each layer from the target model, from top to bottom.
In this fashion, our framework is intended to achieve higher personalization capacity for both simple and complex patterns (\eg, image and label-level personalization). 
We define a uniform personalization partition rate $p\in[0,1]$ to determine the precise proportion of the personalized channels in each layer.
It follows that these $p$ proportion of channel parameters are trained locally, without the aggregation by the central server. 
Subsequently, these private weights vary from each other, written as $w_i$ where the subscript is associated with its client ID. 
The remaining $(1-p)$ percentage of the shared weights, denoted as $w_0$, are trained with common FL algorithms, such as FedAvg~\cite{fedavg}.

A larger value in $p$ represents a higher degree of personalization. Therefore, the case when $p=0$ degenerates into the conventional FedAvg \cite{fedavg} with no model personalization, and conversely $p=1$ denotes a full local training procedure in an absence of the federated communications. 
One significant benefit from our vertical decoupling strategy, compared with horizontally layer-wise personalization ones, such as LG-Fed \cite{lgfed} and Fed-Per \cite{fedper}, is to enable a model personalization from lower to topper layers, resulting in a potential to a more general framework for weights personalization, as well as improving its capacity to address a broader range of data heterogeneity, such as both feature and label distribution heterogeneity.

\para{Progressive model personalization.} One key element in our proposed channel decoupling scheme is the determination of the personalized ratio $p$ in each layer. 
As aforementioned, the personalized ratio $p$ controls the volume of private weights to learn the local representation, which determines the capacity to learn a good representation on the heterogeneous data.
We are motivated to provide a better initialization for the personalization from the globally learned representation.
Thus, as shown in Figure~\ref{fig:kd}, instead of a fixed $p$, the model capability to learn local personalized features is taken into account by a progressive increment scheme. 
That is, in the initial stage, we set $p$ to a small value to facilitate learning global representation for faster convergence, and afterwards gradually increase its value. 
Consequently, we increase the value of $p$ progressively depending on the global epoch number $T$. 
Similar to the learning rate schedule, a variety of schemes exist for the increment, such as cyclical learning rate \cite{smith2017cyclical}, cosine annealing \cite{loshchilov2016sgdr}. 
For simplicity, here we apply the linear growth scheme, \ie,
\begin{equation}
    p_t = p \cdot \frac{t}{T} \label{eq:p},
\end{equation}
where $T$ is the total global epoch number, $t$ is the current epoch number, and $p$ is the maximum personalization ratio.

\subsection{Cyclic Distillation}
The backbone neural network is decoupled into personalized weights $w_i$ and shared weights $w_0$ respectively, and afterwards trained simultaneously by optimizing the local supervised loss in Eq. \eqref{eq:local_loss}. 
However, as the local personalized and global parameters are learned with different distribution data, the statics of these parameters suffer from divergence and subsequently lead to the performance degradation~\cite{fednoniid}.
Additionally, explicit consistency regularization between two parts is absent, during the optimization of local supervised objectives in most previous works. 
To cope with this issue, we make the first attempt at introducing self distillation into \pfl~to improve the client-side weights inner communications between the private and shared model weights and thus reduce the gap between the learned representations from local and global weights.  
Motivated by inplace distillation~\cite{inplacekd}, the key idea in the proposed Cyclic Distillation is to impose a consistency regularization between $w_i$ and $w_0$ in the local training procedure, as depicted in \figref{fig:framework}.

We write the subnet parameterized by $w_i$, $w_0$ as $g_{w_i}$, $g_{w_0}$, and the network composed of $(w_i,w_0)$ as $g_{w_i,w_0}$. 
Notably, $g_{w_i}$ intends to learn personalized local representation from $\mathcal{D}_i$, whereas $g_{w_0}$ to learn global general representation. 
For each input sample $\textbf{x}_i$, we collect the predictions $\widetilde{y}_i$, $\widetilde{y}_i^L$, $\widetilde{y}_i^G$ from $g_{w_i,w_0}$, $g_{w_i}$, $g_{w_0}$, respectively. 
The overall prediction $\widetilde{y}_i$ minimizes the cross entropy loss $\mathcal{L}_{CE}$ with ground truth $y_i$. 
The cyclic distillation loss is defined as:
\begin{equation}
    \mathcal{L}_{CD} = \frac{1}{2}\left(KL(\widetilde{y}_i^L,\widetilde{y}_i^G) + KL(\widetilde{y}_i^G,\widetilde{y}_i^L)\right), \label{eq:kd}
\end{equation}
where $KL(\cdot,\cdot)$ denotes the Kullback-Leibler (KL) divergence. 
It can impose an consistency regularization between $w_i$ and $w_0$, guiding the predictions from $w_i$ and $w_0$ to align with each other. 
Consequently, the overall loss function is 
\begin{equation}
    \mathcal{L} = \mathcal{L}_{CE} + \lambda \cdot \mathcal{L}_{CD}, \label{eq:loss}
\end{equation}
with the balancing coefficient $\lambda$ set to 1 in this work.

\paragraph{Temporal average moving for personalized weights.} To stabilize the training performance, we utilize an exponential moving average (EMA) scheme for the local weights update in personalized channel $w_i$ for a more smoothing training dynamics. 
We use the superscript $l$ to mark the corresponded local epoch number, then the EMA update of $w_i$ at $t$ is 
\begin{equation}
    w_i^l = \beta_t w_i^{\prime l} + (1-\beta_t) w_i^{l-1}, \label{eq:ema}
\end{equation}
where $w_i^{\prime l}$ is the raw update from Eq. \eqref{eq:loss}. The smoothing coefficient $\beta_t$ depends on the current global epoch number and follows a ramp-up strategy in previous works \cite{rampup} \ie,
\begin{equation}
    \beta_t = \begin{cases}
    \beta \cdot \exp(-5(1 - \frac{t}{t_0})^2), & t \le t_0 \\
    \beta, & t > t_0
    \end{cases}
    ,\label{eq:alpha}
\end{equation}
where $\beta$ is set to 0.5, and $t_0$ is set to 10\% of the total federated epoch numbers.

\paragraph{Method overview.} We summarize the thorough local training procedure at each client in Alg. \ref{alg}. Afterwards, the central server collects all global weights $w_0$ from each client and adopts FedAvg to aggregate them. 

\begin{algorithm}[tbp!]
\caption{Local training with \name~at client $i$.} \label{alg}
\begin{algorithmic}[1]
\Require local epoch number $\eta_i$
\Ensure $w_0^t$
\State Download $w_0^{t-1}$ from Central Server 
\State Update Personalized Ratio $p$
\For $~l=1,2,\cdots,\eta_i$
\State Sample of Batch of Data from $\mathcal{D}_i$
\State Forward and Compute Cross Entropy Loss $\mathcal{L}_{CE}$
\State Compute Cyclic Distillation Loss $\mathcal{L}_{CD}$ in Eq. \eqref{eq:kd}
\State Update the Weights
\State Adjust Personalized Weights $w_i^l$ with Eq. \eqref{eq:ema} 
\EndFor
\State Upload $w_0^t$ to Central Server
\end{algorithmic}
\end{algorithm}

\section{Experiments}

\subsection{Datasets}
Focusing on image classification tasks, we use four benchmark datasets to evaluate the proposed \name, namely CIFAR-10, CIFAR-100, FLICKR-AES, and a combination of public and private histology images, termed as HISTO-FED in this paper. 

\para{CIFAR-10} contains a total number of 60000 color images sized at $32\times32$ in 10 classes, with 5000 training images and 1000 test images per class \cite{cifar}. 
We focus on a highly non-IID setting, \ie characterized as label distribution skew. We follow previous works~\cite{lgfed,fedavg} to assign images from at most $s \in \{2,3,4,5,8,10\}$ classes to each client. A higher $s$ corresponds to higher variance in data distribution. For example, $s=10$ is an IID setting, while $s=2$ is the highest heterogeneous data split. We set the client number $K=10$ for CIFAR-10, as the literature \cite{lgfed,fedper}. 

\para{CIFAR-100} contains 500 training images and 100 testing images per class, with a total number of 100 classes \cite{cifar}. 
Similar to CIFAR-10, the color images are scaled at $32\times32$. We set the client number $K=30$ and assign at most $s=40$ classes to each client \cite{fedper}, which is also non-IID (\ie label distribution skew).

\para{FLICKR-AES} is used to evaluate the performance of personalized image aesthetics in many literature \cite{pia}.
%
%
%
The images are randomly split to 80\% for training and 20\% for testing. 
Additionally, REAL-CUR is leveraged as an external test set to evaluate the global model representation in the context of real-world personal photo ranking. 
Images from 14 personal albums, with an average of 197 to 222 images per album, were collected and rated by one user \cite{pia}. 
Due to the personal bias in aesthetic scoring, the non-IID is characterized as concept shift, leading to non-IID data distribution. We use a subset of $K= 30$ users as clients, the same as the setting in previous work \cite{fedper}.

\begin{figure}[h!]
\centering
\includegraphics[width=0.8\linewidth]{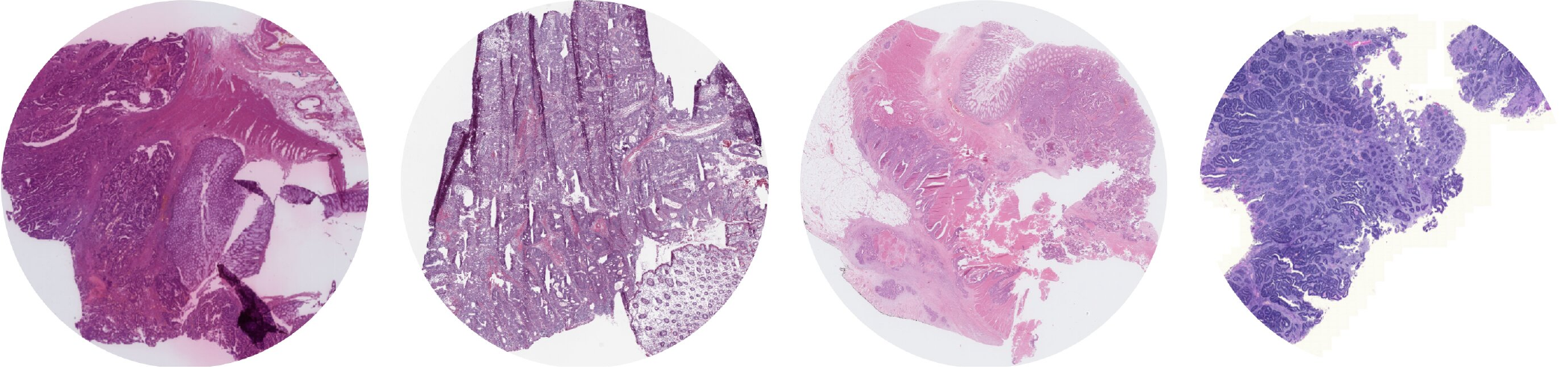}
\caption{Illustrate examples of stain variant histology whole slides from different medical centers.}
\label{fig:data}
\end{figure}

\para{HISTO-FED} is a medical image datasets, consisting of both public and private hematoxylin \& eosin (H\&E) stained histological whole-slide images of human colorectal cancer (CRC) and normal tissue. 
They are curated from four medical centers. We set the client number $K=3$ where each of them uses images from one medical center, and the remaining center is used as the external test set. Client 1 and client 2 use subsets of total slides number N = 86 and 50 from two public datasets NCT-CRC-HE-100K, CRC-VAL-HE-7K respectively.
Each of them has a number of 7180 image patches, spitted from slides. Client 3 and external test set have 7000, and 4000 image patches respectively, curated from a private dataset of slides number $N=20$ and 10. Each image is labeled with one of nine categories. All images involved in this research received appropriate ethical approval. Due to the stain variance \cite{histofed}, the images between clients are highly non-IID, depicted as the feature skew. (as illustrated in \figref{fig:data}). 

\subsection{Experimental Settings}

\para{Backbone Architectures.} We adopt the following network backbones for performance evaluation: 1) LeNet-5 \cite{lenet} for CIFAR-10, 2) ResNet-34 \cite{resnet} for CIFAR-100, FLICK-AES, and 3) ResNet-32 for HISTO-FED, following the previous works \cite{lgfed,fedper,histofed}. All backbone networks are trained from scratch, without loading any pre-trained weights.

\para{Hyper Parameters.} 
%
At each local client, we employ stochastic gradient descent optimizer where the Nesterov momentum and the weight decay rate are set to 0.9 and $5\times10^{-4}$ respectively. 
The local epoch number $\eta_i=1$, batch size $b=128$ for CIFAR-10; $\eta_i=4$, $b=128$ for CIFAR-100; $\eta_i =4$ $b=4$ for FLICK-AES and HISTO-FED. The total epoch number $T$ is set to 50.
Additionally, in \name, we set maximum smoothing coefficient for EMA $\alpha=0.5$ in Eq. \eqref{eq:alpha}, balancing coefficient in loss function $\lambda=1$ in Eq. \eqref{eq:loss}, and $p=0.5$ on CIFAR-10/100 and HISTO-FED and $p=0.8$ on FLICK-AES in Eq. \eqref{eq:p}.

\para{Comparison Methods.} We compare our \name~ with FedAvg \cite{fedavg}, local training, LG-Fed \cite{lgfed} and FedPer \cite{fedper}. FedAvg is the conventional federated learning algorithm, where no personalization is involved \cite{fedavg}. Local training trains a collection of $K$ models for each client, without communications between clients. Personalized Federated Learning models play an intermediate role between FedAvg and local training, by personalizing the global structure with local unshared weights. In this work, we primarily compared our method with a model modification-based personalization scheme, which is more similar to ours. LG-Fed jointly learns compact local representations on each device with lower layers and a global model in top layers across all clients \cite{lgfed}. FedPer designed a base plus top personalization layer structure, conversely to LG-Fed which assigns personalization to bottom layers.

\para{Implementations.} All experiments are conducted on one NVIDIA Tesla V100 GPU with 32Gb memory. The proposed \name ~is implemented on Pytorch 1.6.0 in Python 3.7.0 environment. We used an public implementation for FedAvg, LG-Fed, and FedPer for comparison. All the extra hyper-parameters involved in the compared methods are retained as their original settings. 
 
\para{Evaluation Metrics.} We use two metrics on CIFAR-10, and CIFAR-100, following previous work \cite{lgfed}. 1) \emph{Local Test Top-1 Classification Accuracy} (\%). We know precisely the client where the data sample belongs, thus we can choose the particular trained local model to predict. It evaluates the performance of model personalization. 2) \emph{New Test Top-1 Classification Accuracy} (\%). We do not know the client where the data sample belongs to, thus we employ an ensemble of all local models to derive averaged predictions, where the local model will be uploaded to the central server. This index measures the compatibility between local and global model representation. On FLICKR-AES and HISTO-FED, we use one more index, namely the  \emph{External Test Top-1 Classification Accuracy} (\%). Specifically, we use external test samples in addition to the local or new test. Thus, these images potentially are potentially sampled from different distributions, intended to verify the generalization on the global model representation. Notably, we do not perform the external validation on CIFAR-10/100 due to the absence of external samples.

\subsection{Experimental Results on Synthesized Data}

\begin{figure}[htbp]
\centering
\includegraphics[width=0.7\linewidth]{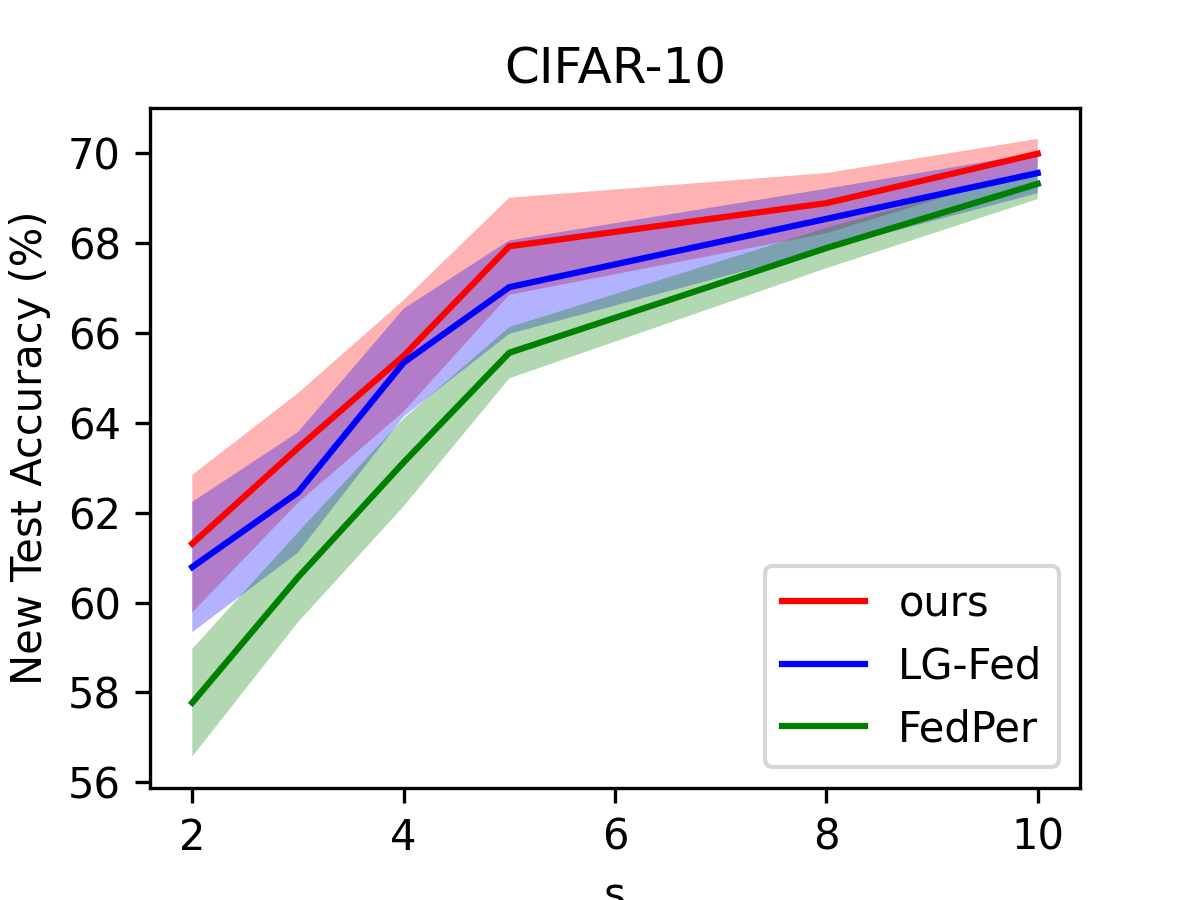}
\caption{Effect of data heterogeneity on model personalization.}
\label{fig:res_data}
\end{figure}

\para{Effect of Data Heterogeneity.} We first evaluate the performance on CIFAR-10 on different levels of data heterogeneity, which is quantified by $s$. As shown in \figref{fig:res_data}, on all degree of heterogeneity, \ie, $s$, \name~consistently outperforms LG-Fed and FedPer. Moreover, the performance gap monotonically increases with the heterogeneity. When $s=10$, \ie, in a IID setting, achieves very marginally the same test accuracy. In the rest of this section, we focus on the most non-IID settings. 

\para{Results on CIFAR-10.} As shown in \Cref{tab-cifar}, our proposed PFL frameworks significantly improve the backbone network by 31.83\%, on the highest degree of data heterogeneity, \ie, $s=2$. 
This empirical success shows the effectiveness of personalization by the channel-wise ensemble. 
Additionally, compared with the state-of-the-art layer-wise personalization scheme \cite{lgfed,fedper}, our approach achieves both the best local and new classification accuracy, suggesting our model learns more capable local and global representation. Additionally, the superiority of the new test accuracy suggests that our scheme also achieves higher generalization on unseen data in personalization, attributed to the equal role of personalized and shared weights they played in FL.

\begin{table}[h!]
\caption{Comparison of personalized federated learning methods on CIFAR-10 with a highest heterogeneity non-IID split , \ie, $s=2$. The best results are marked in \textbf{bold}, and results reported in \cite{lgfed} are indicated by *. Two metrics namely the local test and new test classification accuracy, are used for evaluating the model personalization and generalization respectively. }\label{tab-cifar}
\centering
\begin{tabular}{l|c|c}
\hline
Methods & Local ($\uparrow$) & New ($\uparrow$)\\
\hline
FedAvg\cite{fedavg}* & 58.99$\pm$1.50 & 58.99$\pm$1.50 \\
 Local Train*& 87.92$\pm$2.14 & 10.03$\pm$0.06 \\
 LG-Fed\cite{lgfed}* & 91.77$\pm$0.56 & 60.79$\pm$1.45 \\
 FedPer\cite{fedper} & 83.29$\pm$0.98 & 57.77$\pm$1.98 \\
 Ours & \textbf{91.82$\pm$0.43} & \textbf{61.31$\pm$1.53} \\
\hline
\end{tabular}
\end{table}

\para{Results on CIFAR-100.} As illustrated in \Cref{tab-cifar100}, a local test accuracy improvement of 28.75\% is achieved with \name, showing its effectiveness on model personalization for the local dataset with more wealthy categories. Meanwhile, there is a 5.92 new test accuracy improvement, yielding its generalization on unseen data. We also outperform layer-wise personalization methods such as FedPer, LG-Fed. 

\begin{table}[h!]
\caption{Comparison of personalized federated learning methods on CIFAR-100 with non-IID split $s=40$. The best results are marked in \textbf{bold}.}\label{tab-cifar100}
\centering
\begin{tabular}{l|c|c}
\hline
 Methods & Local ($\uparrow$) & New ($\uparrow$)\\
\hline
 FedAvg\cite{fedavg} & 29.23$\pm$1.75 & 29.23$\pm$1.75 \\
 Local Train& 44.59$\pm$0.90 & 11.98$\pm$0.22 \\
 LG-Fed\cite{lgfed} & 56.77$\pm$0.75 & 34.50$\pm$1.02 \\
 FedPer\cite{fedper} & 53.24$\pm$2.33 & 30.47$\pm$1.73\\
 Ours & \textbf{57.98$\pm$0.64} & \textbf{35.15$\pm$0.56} \\
\hline
\end{tabular}
\end{table}

\subsection{Experimental Results on Real-world Data}

\para{Results on FLICKR-AES.} We test the local training performance on FLICKR-AES due to the small scale of local clients, making it easy to suffer from overfitting. 
In, FLICR-AES, the labeling distribution is non-IID, fitting the philosophy of FedPer. With a marginal outperform to LG-FED, FedPer shows its superiority of top layer personalization in tackling with label distribution skew, while LG-Fed suffers an inferior performance. It is worth noticing that LG-Fed only slightly outperforms the baseline FedAvg, this is attributed to the fact that skew exists in label distributions where the personalized bottom layers in LG-Fed are difficult to learn. The empirical comparison is summarized in \Cref{tab-aes}, where our framework outperforms state-of-the-art personalization schemes on both local and external tests. Additionally, \name~can significantly reduce the test variance, leading to more stable and robust predictions. 
Conclusively, \name~equipped with both top and bottom personalization does not suffer from the ill-effect on LG-Fed in facing label distribution skew.

\begin{table}[h!]
\caption{Comparison of personalized federated learning methods on FLICKR-AES, and the external validation of REAL-CUR. The best results are marked in \textbf{bold}.}\label{tab-aes}
\centering
\begin{tabular}{l|c|c}
\hline
 Methods & Local ($\uparrow$) & External ($\uparrow$)  \\
\hline
 FedAvg\cite{fedavg} & 24.50$\pm$2.01 & 20.08$\pm$1.34 \\
 LG-Fed\cite{lgfed} & 25.78$\pm$2.40 & 20.98$\pm$1.34 \\
 FedPer\cite{fedper} & 43.26$\pm$3.23 & 40.55$\pm$1.78 \\
 Ours & \textbf{47.89$\pm$2.03} & \textbf{45.67$\pm$1.67} \\
\hline
\end{tabular}
\end{table}

\para{Results on HISTO-FED.} The internal and external test results of \name~ on four clients consistently outperform the baselines, in \Cref{tab-med}. We achieve higher improvement from the internal validation than the external one, which suggests that our model can personalize the global model well. These empirical results show the robustness and success of federated personalization of \name~ on medical images, in addition to natural image classification.

\begin{table}[h!]
\caption{Results of real-world medical images, \ie, stain variant histology slides, on each clients. Performance are evaluated by client side test accuracy. The best results are marked in \textbf{bold}.}\label{tab-med}
\centering
\resizebox{1\linewidth}{!}{
\begin{tabular}{l|c|c|c|c}
\hline
Methods & client \#1 & client \#2 & client \#3 & External ($\uparrow$)\\
\hline
FedAvg\cite{fedavg} & 65.23 & 65.31 & 65.45 & 60.03\\
Local Train& 75.53 & 74.87 & 74.21 & 34.31 \\
LG-Fed\cite{lgfed} & 76.32 & 76.90 & 77.01 & 63.22\\
FedPer\cite{fedper} & 75.43 & 75.21 & 75.56 & 57.89\\
Ours & \textbf{77.39} & \textbf{77.45} & \textbf{77.38} & \textbf{65.66}\\ 
\hline
\end{tabular}
}
\end{table}

\para{Discussion.} Comprehensive experiments on four datasets, characterized with different non-IID settings, confirmed that our \name~is the only method to consistently achieve state-of-the-art results.
Although, LG-Fed performs better to FedAvg in the existence of feature skew, and FedPer better with a label distribution skew, their performance sharply declines when the non-IID settings are interchanged.  
It is assumed that LG-Fed personalizes the bottom layer to better learn from highly heterogeneous images, while FedPer personalizes the top layer to distinguish unbalanced samples. 
Our \name, containing both the low-level and high-level personalization can reduce the reliance on prior knowledge for personalization decisions. 

\subsection{Ablation Analysis}
\begin{figure}[htbp]
\centering
\includegraphics[width=0.63\linewidth]{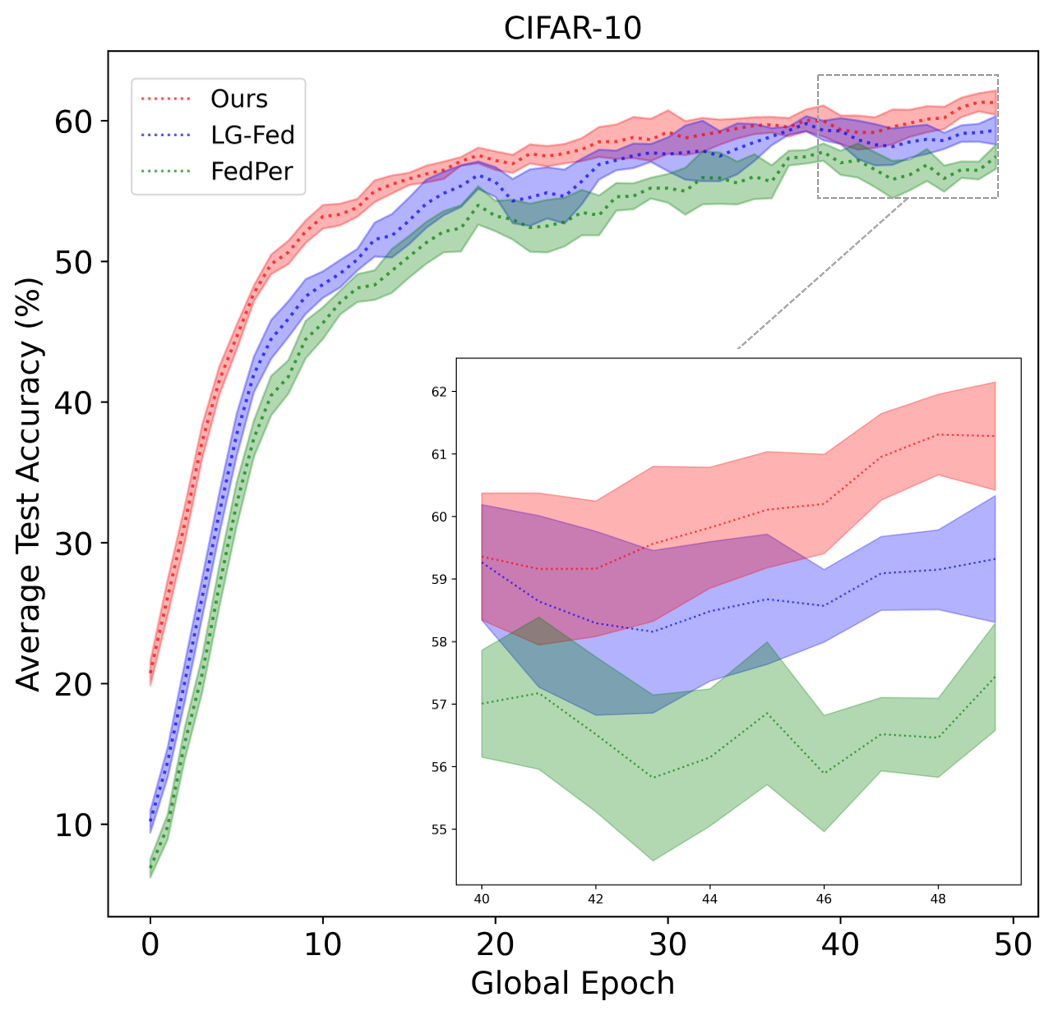}
\caption{Effect on the performance of LeNet-5 on CIFAR-10, compared with different PFL frameworks \cite{lgfed,fedper}. As illustrated, our \name~achieve higher test accuracy, together with a significant faster convergence speed.}
\label{fig:res_cifar}
\end{figure}

The proposed \name~is composed of three functional components to assist the channel decoupling, namely the progressive personalization ratio increment scheme (LI), temporal average moving for the personalized weights (TA), and cyclic distillation (CD). To test the effectiveness of each scheme, we perform ablation studies on CIFAR-10 with a $s=2$ split. As shown in \Cref{tab:abl}, we can observe that 1) with all components, \name~achieves the best performance, demonstrating the effectiveness of integrating three schemes \ie LI + TA + CD, with a 1.51\% local test accuracy improvement to the origin channel decoupling; 2) CD achieves the highest improvement, TA second, and LI the least; 3) LI and TA can stabilize the training, resulting to a smaller stand deviation. We also visualize the training performance of non-IID CIFAR-10 in \figref{fig:res_cifar}, where our \name~ achieves a faster convergence compared with existing layer-wise personalization methods, which requires fewer communication rounds during the federation.

\begin{table}[htbp!]
\caption{Ablation study on non-IID CIFAR-10 split, with $s=2$, to evaluate the effectiveness of each component.}
\label{tab:abl}
\centering
\begin{tabular}{c|c|c|c|c}
\hline
LI & TA & CD & Local ($\uparrow$) & New ($\uparrow$) \\ 
\hline
&&& 90.31$\pm$0.67 & 59.12$\pm$0.32 \\
\cmark&&& 90.36$\pm$0.65 & 59.14$\pm$0.30 \\
&\cmark&& 90.45$\pm$0.21 & 59.45$\pm$0.54 \\
&&\cmark & 90.58$\pm$0.44 & 60.57$\pm$0.32 \\
&\cmark&\cmark & 91.67$\pm$0.54 & 61.20$\pm$2.03 \\
\cmark& &\cmark & 91.00$\pm$1.03 & 59.84$\pm$1.53 \\
\cmark&\cmark& & 90.81$\pm$0.38 & 59.34$\pm$0.56 \\ 
\cmark & \cmark & \cmark & \textbf{91.82$\pm$0.43} & \textbf{61.31$\pm$1.53} \\
\hline
\end{tabular}
\end{table}

\section{Limitations and Conclusions}
In this paper, we propose \name~to vertically decouple channels in the global model for personalization. 
Our vertical decoupling method can personalize the local model, with guidance towards learning on high- and low-level feature representation. 
Subsequently, it can handle a variety of settings of data heterogeneity including the feature skew, label distribution skew, and concept shift. 
Empirically, compared with the previous layer-wise split which only learns one part of them, our framework shows a consistent success on four benchmark datasets. 
We also propose cyclic distillation to impose a consistency regularization and prevent the weights divergence in personalization.
However, the cyclic distillation is currently designed by using soft predictions, which restricted itself to classification tasks. The extensions to segmentation and detection are left to future work.
To stabilize the training process, we leverage a temporal average moving for personalized weights and a progressive increase scheme for the personalization ratio. 
Yet, we primarily assign a fixed personalization ratio for all layers, which yields an interesting future direction on searching a layer-specific optimal ratio.

\paragraph{Acknowledgements.} The work described in this paper is supported by grants from HKU Startup Fund and HKU Seed Fund for Basic Research (Project No. 202009185079).

{\small
\bibliographystyle{ieee_fullname}
\bibliography{egbib}
}

\end{document}